\documentclass[runningheads]{llncs}
\usepackage{float}
\usepackage{graphicx}
\usepackage{amsmath,amssymb}%,amsthm}
\usepackage{etoolbox}
\usepackage{todonotes}
\usepackage[ruled,noend,linesnumbered,noline]{algorithm2e}
\SetFuncSty{textsc}
\usepackage{hyperref}
\usepackage[noabbrev,capitalise]{cleveref}
\usepackage{array}
\usepackage{booktabs}
\usepackage{tabularx}
\usepackage{makecell}
\usepackage{multirow}
\usepackage{multicol}
\usepackage{csquotes}
\usepackage{marvosym}
\usepackage{orcidlink}
\usepackage{siunitx}
\usepackage{xfrac}

\newcolumntype{P}[1]{>{\centering\arraybackslash}p{#1}}
\makeatletter
\patchcmd{\@algocf@start}% <cmd>
  {-1.5em}% <search>
  {0pt}% <replace>
  {}{}% <success><failure>
\makeatother

\newcommand{\Markline}[1]{\leavevmode\rlap{\hbox to \hsize{\color{#1}\leaders\hrule height .8\baselineskip depth .5ex\hfill}}}

\begin{document}
%
%\title{Model-based Evolutionary Constant Optimization in GP-GOMEA for Symbolic Regression}%\thanks{Supported by organization x.}}
\title{Simultaneous Model-Based Evolution of Constants and Expression Structure in GP-GOMEA for Symbolic Regression}
\titlerunning{Model-based Evolutionary Constant Optimization in GP-GOMEA}
% If the paper title is too long for the running head, you can set
% an abbreviated paper title here
%
% \orcidID{0000-1111-2222-3333}
\author{Johannes Koch\inst{1,2}${}^{\text{(\Letter)}}$\orcidlink{0009-0008-7570-7621}\and
Tanja Alderliesten\inst{3}\orcidlink{0000-0003-4261-7511} \and
Peter A.N. Bosman\inst{1,2}\orcidlink{0000-0002-4186-6666}}
\authorrunning{J. Koch et al.}
% First names are abbreviated in the running head.
% If there are more than two authors, 'et al.' is used.
%
\institute{Centrum Wiskunde \& Informatica, Amsterdam, The Netherlands\\
\email{\{Johannes,Peter.Bosman\}@cwi.nl} \and
Delft University of Technology, Delft, The Netherlands \and
Leiden University Medical Center, Leiden, The Netherlands \\
\email{T.Alderliesten@lumc.nl}}
\maketitle              % typeset the header of the contribution
\begin{abstract}
%\todo{Please check the authors, they are copied from Joe's last ppsn paper}{}
Genetic programming (GP) approaches are among the state-of-the-art for symbolic regression, the task of constructing symbolic expressions that fit well with data. To find highly accurate symbolic expressions, both the expression structure and any contained real-valued constants, are important. GP-GOMEA, a modern model-based evolutionary algorithm, is one of the leading algorithms for finding accurate, yet compact expressions. Yet, GP-GOMEA does not perform dedicated constant optimization, but rather uses ephemeral random constants. Hence, the accuracy of GP-GOMEA may well still be improved upon by the incorporation of a constant optimization mechanism. Existing research into mixed discrete-continuous optimization with EAs has shown that a simultaneous and well-integrated approach to optimizing both discrete and continuous parts, leads to the best results on a variety of problems, especially when there are interactions between these parts. In this paper, we therefore propose a novel approach where constants in expressions are optimized at the same time as the expression structure by merging the real-valued variant of GOMEA with GP-GOMEA. The proposed approach is compared to other forms of handling constants in GP-GOMEA, and in the context of other commonly used techniques such as linear scaling, restarts, and constant tuning after GP optimization. Our results indicate that our novel approach generally performs best and confirms the importance of simultaneous constant optimization \emph{during} evolution.

\keywords{Genetic programming \and Constant optimization \and Symbolic regression \and Model-based evolutionary algorithms.}
\end{abstract}

\section{Introduction}
In recent years, the field of eXplainable AI (XAI) has received increased attention, especially for use cases where AI models can affect lives and livelihoods. Given a dataset and a library of atomic functions such as \(\{+,-,\times,\div,\sin\}\), symbolic regression (SR) is the task of finding an interpretable expression that best describes the relation between one (output) variable and other (input) variables~\cite{Koza1994}. Compact SR models (i.e., expressions) are interesting from the perspective of XAI and interpretable ML (IML), as they are readable and therefore have the potential to be humanly understandable~\cite{Rudin2019,GPforIML2022}.
GP-GOMEA is a genetic programming (GP) based SR method that is amongst the state-of-the-art for finding compact, yet accurate expressions and part of the non-dominated front on the recent SR benchmark SRBench~\cite{SRBench2021,virgolin2021improving}.

GP approaches generally primarily optimize the expression structure through a process of iteratively recombining individuals in a population of (initially random) expressions. The strength of GP-GOMEA in particular is finding expression structures by dynamically learning and exploiting a linkage model during optimization that captures key dependencies between parts of an expression template. Still, in general, for an SR expression to be highly accurate, not only must the right expression structure be found, also the real-valued coefficients must be optimized. Typically, ephemeral random constants (ERCs) are used~\cite{Koza1994}, which are random constants that are sampled during initialization and then not modified any further. Current approaches refine these by performing gradient-based search or randomly mutating constant values. While the expression structure is optimized by performing variation on the whole population, constant optimization typically is performed on individual solutions. %Constant optimization still is an active topic of research and is considered to be an open issue in GP~\cite{ONeill2010}. 
In the context of XAI, where SR expressions need to be compact, finding better constants is expected to increase expression accuracy while keeping expressions compact. %Typically, random constants are sampled during initialization and then refined by performing gradient-based search or randomly mutating the coefficient values. Where the expression structure is optimized across the whole population, constant optimization typically is performed on single solutions.

In this paper, we present and evaluate GP-RV-GOMEA, a new approach to constant optimization in GP-GOMEA that takes inspiration from GAMBIT~\cite{Sadowski2018}, a fully integrated model-based evolutionary algorithm (MBEA) approach to mixed discrete and continuous optimization. To make constant optimization a first-class citizen in GP-GOMEA, SR is considered to be a mixed discrete and real-valued problem, where GP-GOMEA is used to optimize the structure of expressions and the real-valued GOMEA (RV-GOMEA) is used to simultaneously optimize the constants of all expressions. 
Compared to random coefficient mutations, the design and use of a dedicated MBEA is expected to lead to a more directed and effective search that is better positioned to overcome potentially ill-conditioned gradients that non-linear-least-squares methods can encounter~\cite{Kronberger2022Local}.

%\todo{Still missing: Glimpse of results? Explicit listing of contributions made?}{}
The remainder of this paper is organized as follows. Related works are first discussed in the following Section. In~\cref{sec:method} we describe the new method GP-RV-GOMEA.  In~\cref{sec:experiments}, we perform experiments to assess the performance of GP-RV-GOMEA as well as other GP-GOMEA variants. We discuss our main findings in~\cref{sec:discussion} and draw our final conclusions in~\cref{sec:conclusion}.

\section{Related Work\label{sec:related-work}}
% Ephemeral random constants (ERCs)~\cite{Koza1994} are random constants that are sampled during initialization and then not modified any further. Since their 
Since the introduction of ERCs, constant optimization in GP has become a well-studied subject and various approaches have been suggested, including coefficient mutations~\cite{CoefficientMutatioVirgolin2022}, gradient-based search~\cite{Harrison2023,Kommenda2019,Rockett2021}, %scaling techniques~\cite{Keijzer2004,Dick2020,Scaling2023} 
and meta-heuristic optimization~\cite{Cerny2008,Howard1995,Mukherjee2012,Cesar2009,Sharman1995}.

In~\cite{Cesar2009} and~\cite{Mukherjee2012}, real-valued EAs are nested within a GP algorithm to separately optimize the constants of either the best or all individuals, respectively. In~\cite{Sharman1995}, simulated annealing is used in the same manner. Coefficient mutations, gradient-based search, and the approaches presented in~\cite{Mukherjee2012,Sharman1995,Cesar2009}, all optimize the constants in expressions separately per expression, either in a nested loop inside GP, or after GP terminates. In this paper, we consider \emph{simultaneous} evolutionary optimization. %\todo{Reword this, the point is that the constants of other individuals in the population are not noise, but likely a useful prior...}{
The motivation for joint optimization is that as the GP population converges over time, similar constant values are likely needed across the population, and thus constant optimization is not viewed as an independent problem for each individual in this work. Moreover, in related work with mixed discrete and continuous variables, their joint optimization has been shown to be advantageous over independent nested, or sequential optimization~\cite{Sadowski2018}.%}

In~\cite{Cerny2008}, differential evolution
%, a real-valued evolutionary algorithm, 
is used to optimize the expression structure and constants at once. However, GP is modeled as a fully real-valued optimization problem by using a fixed number of decision variables to encode the expression structure, combined with a mapping from continuous values to GP operators. The remaining decision variables are the constant values available to an individual. Instead of ERCs, constant references are added that are substituted with the corresponding constant value during evaluation. Our method differs from this approach by interleaving optimization of the structure and constants, and using separate algorithms to do so. A similar approach has been presented in~\cite{Howard1995}, however, significant progress both in GP and real-valued optimization has been made since 1995. Moreover, our method further includes an optimization to avoid unnecessary fitness evaluations, not present in any of the previous approaches. %\todo{As in~\cite{Howard1995} and~\cite{Cerny2008} use very similar setups with constant references, but they don't compare to other methods.}{
To the best of our knowledge, this also is the first work comparing such a form of constant optimization with other approaches.%}

An approach to mixed discrete and real-valued optimization using GOMEA has been presented in~\cite{Sadowski2018}. Our new method takes direct inspiration from that work but uses GP-GOMEA as a specialized GP algorithm for the discrete part and RV-GOMEA for the constants. Regarding constants in GP-GOMEA, ERCs and coefficient mutation have been explored in~\cite{virgolin2021improving} and~\cite{CoefficientMutatioVirgolin2022} respectively. Both approaches are used as a baseline for the new method we introduce here.

\section{Model-based Evolutionary Constant Optimization in GP-GOMEA\label{sec:method}}

In this section, GP-RV-GOMEA, a combination of GP-GOMEA and RV-GOMEA based on GAMBIT is presented. First, the family of gene-pool optimal mixing evolutionary algorithms (GOMEAs) and the individual algorithms used are shortly described before the combination is introduced.

\subsection{Gene-pool Optimal Mixing Evolutionary Algorithms}

Similar to other population-based algorithms, GOMEAs work by iteratively refining an initially random set of candidate solutions. Each individual in a GOMEA is typically represented as a fixed-length list of decision variables. To achieve effective optimization, the aim in a GOMEA is to model and exploit dependencies between linked variables~\cite{Arkadiy2021}. % produced capable algorithms for various domains, including GP~\cite{Arkadiy2021,Bouter17MO,virgolin2021improving,Sadowski2018}.

These dependencies can be set a priori or learned in every generation during optimization. This linkage information is modeled using a so-called family of subsets (FOS) structure, a set that contains subsets of all decision variable indices. Each subset in the FOS then corresponds to a group of linked variables and is used as a crossover mask during variation. To learn the FOS during optimization, hierarchical clustering using UPGMA~\cite{Gronau2007} is often used on the decision variables. As a similarity measure, typically the mutual information between the decision variables is used~\cite{Arkadiy2021}. Starting from all single variable subsets, a subset is added for every pair of joined subsets during hierarchical clustering. This results in a so-called Linkage Tree FOS.

Variation then is performed for each individual subset in the FOS, where all variables in the given subset are varied together. For discrete variables, the new values are inherited from a donor randomly picked from the population, or sampled from a previously estimated multivariate normal distribution in the real-valued case. The changed solution is then evaluated, and reverted in case the fitness is worse compared to before the modification. This variation procedure is called Gene-pool Optimal Mixing (GOM) and is performed for all FOS subsets and individuals in a single generation. For discrete decision variables, the donors are a copy of the population that is not modified, to match the distribution of values at the time of learning the linkage model. The real-valued distribution used corresponds to a potentially adapted (see~\cref{RVGOMEAsubsection}) maximum-likelihood estimate of the top $35\%$ solutions in the population for every FOS subset.

Further, forced improvements, a mechanism which forces solutions that did not improve within the last \(1+\log_{10}(\text{population size})\) generations, are used~\cite{Arkadiy2021}. This mechanism subjects such solutions to an additional round of GOM steps until the solution has improved, where the donor is the elite of the current population. If no improvement can be found after processing all FOS subsets, the solution is replaced with the elite solution.%
\iffalse
\vspace{-0.5cm}
\input{code/gom}
\vspace{-1cm}
\fi

\subsection{GP-GOMEA}

GP-GOMEA~\cite{virgolin2021improving} is the GP variant of GOMEA, where a fixed, tree-based symbolic expression template is mapped to a string representation. This inherently limits the maximum size of the learned expressions. Typically, full \(n\)-ary trees are used, where \(n\) is the largest arity in the function set used. Using a fixed template introduces syntactic introns for expressions smaller than the template size, i.e., all subtrees of terminal nodes such as input features or constants have no impact on the symbolic expression encoded by a solution. This is exploited during GOM when no actively used node is changed. The inherited modifications are simply accepted as the fitness remains unchanged.

\subsection{RV-GOMEA\label{RVGOMEAsubsection}}

RV-GOMEA~\cite{Bouter17SO} is the real-valued version of GOMEA for continuous search spaces.% Compared to the discrete and GP version, during GOM the changed values are not inherited from another solution in the population but sampled from a multivariate normal distribution. This distribution corresponds to the maximum likelihood estimate of the top $35\%$ solutions in the population for every FOS subset.
When the linkage is not set a priori, then the similarity metric used during clustering typically is the Pearson product-moment correlation coefficient. However, in this paper, only the full FOS, i.e., a single crossover mask containing all variables, is used as all constants used in a solution can affect each other.

In addition to sampling new values from a learned distribution, additional mechanisms detailed in~\cite{Bouter17SO} influence the variation compared to other GOMEA variants. These are the Anticipated Mean Shift (AMS), which shifts a part of the population in the direction of the mean shift between the previous and current generation, and Adaptive Variance Scaling (AVS) to adapt the step size in case solutions are found more than a standard deviation away from the distribution mean. Forced improvements in the real-valued case are performed by bisecting the values of the solution and the elite for each FOS subset.

\subsection{GP-RV-GOMEA}

\subsubsection{Solution Representation} 
\begin{figure}[htbp]
    \centering
    \vspace{-0.45cm}
    \includegraphics[width=0.95\linewidth]{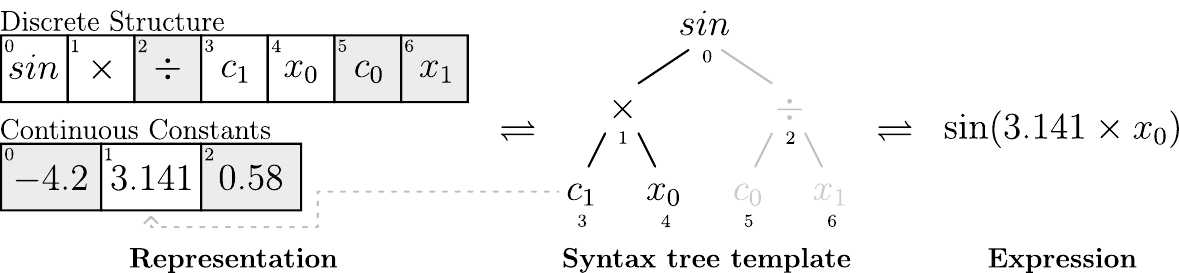}
    \vspace{-0.25cm}
    \caption{The genotype (left) of a single individual in GP-RV-GOMEA and how it relates to the encoded expression (right). Shaded values are introns that do not affect the semantic meaning of the expression.\label{fig:genotype}}
    \vspace{-1.3cm}
\end{figure}

To make constant optimization a first-class citizen, the GP-GOMEA genotype consisting of discrete decision variables is extended with a fixed number of real-valued constants that will be optimized using RV-GOMEA. Instead of special constant nodes, the GP terminal set is extended with constant references for all added constants. This representation is shown in~\cref{fig:genotype} and during evaluation, constant references are substituted with the corresponding value from the real-valued decision variables. Note that introns can be both discrete or real-valued.

\subsubsection{Interleaving Scheme} Using this mixed representation, optimization then follows the approach described in~\cite{Sadowski2018} with additional modifications specific to GP. The approach is outlined in~\cref{code:rvgpg}. After initialization, in every generation, first, the discrete linkage model used by GP-GOMEA is learned from the current population. This is followed by a mixed variation procedure that interleaves performing discrete GOM using GP-GOMEA and real-valued variation steps using RV-GOMEA until all subsets in the discrete FOS have been processed. In~\cite{Sadowski2018}, this interleaving is done by shuffling the order of all variation steps, both discrete and continuous, to balance the computational effort spent. However, as GP-GOMEA only performs evaluations if actively used nodes are modified, this quickly leads to an imbalanced distribution of computational effort between the discrete and real-valued optimization. Hence, a different approach directly based on the number of evaluations performed for each variable type is used. To balance the computational effort spent, the ratio of real-valued to discrete evaluations  (\(\in [0,\infty]\)) is transformed into a probability of performing a real-valued step and then the next step is sampled from the uniform distribution (lines 6-7 in~\cref{code:rvgpg})%\footnote{For \(\frac{\text{\#RV evaluations}}{\text{\#GP evaluations}}>1\) the sigmoid function would lead to the "correct" balancing as a ratio of \(\geq 2\) leads to \(p_{RV} \leq 0\), but the used transformation ensures that the imbalance towards performing more real-valued steps is bounded.}%
. Note that while GP steps amount to performing GOM for all solutions and a single FOS subset, the real-valued steps amount to performing a full RV-GOMEA generation. Thus, the real-valued steps consist of computing the maximum-likelihood estimate, performing GOM, and finally AMS.

\vspace{-0.6cm}
\begin{algorithm}
\DontPrintSemicolon

\SetKwInOut{Parameters}{Parameters}
\SetKwInOut{Variables}{Variables}

% \Parameters{\;
% \Indp
% Population size $N$\;
% Target RV to GP ratio $\theta_{RV:GP}$
% }

\SetKwFunction{RandomSolution}{RandomSolution}
\SetKwFunction{Evaluate}{Evaluate}
\SetKwFunction{UpdateArchive}{UpdateElitistArchive}
\SetKwFunction{UpdateArchive}{UpdateElitistSolution}
\SetKwFunction{RVGOMEA}{RVGOMEA}
\SetKwFunction{ShouldTerminate}{TerminationCriteriaSatisfied}
\SetKwFunction{PerformConstantRenaming}{PerformConstantRenaming}
\SetKwFunction{ClusteringAndRegistration}{ClusteringAndRegistration}

\SetKwFunction{LearnLinkageModel}{LearnLinkageModel}
\SetKwFunction{TakeRandom}{TakeRandom}
\SetKwFunction{GPStep}{GPStep}
\SetKwFunction{RVStep}{RVStep}
\SetKwFunction{GPFI}{GPForcedImprovements}
\SetKwFunction{InitRV}{InitializeRVStates}
\SetKwFunction{GetClusterDonors}{GetClusterDonors}
\SetKwFunction{RVToGPEvaluationRatio}{RVToGPEvaluationRatio}
\SetKwFunction{ComputeMR}{ComputeConstantMismatches}
\SetKwFunction{ApplyForcedImprovements}{ApplyForcedImprovements}
\SetKwFunction{InitializeAndEvaluatePopulation}{InitializeAndEvaluatePopulation}

%\BlankLine
% $\mathcal{P} \longleftarrow \{$ \RandomSolution{} $|\ i\in [0,N-1]\ \}$\;
$\mathcal{P} \longleftarrow$ \InitializeAndEvaluatePopulation{}\;
% \Evaluate{$\mathcal{P}$}\;
%$\mathcal{E} \longleftarrow$ \UpdateArchive{$\mathcal{P}$}\;
% $\mathcal{R} \longleftarrow$ \InitRV{$\mathcal{P}$}\;
\While{$\lnot$\ShouldTerminate}{
    $\mathcal{O} \longleftarrow \mathcal{P}$\;
    $\mathcal{F} \longleftarrow$ \LearnLinkageModel{$\mathcal{P}$}\;
    \While{$\mathcal{F}$ not empty}{
        % $r_{RV:GP} \longleftarrow \frac{\text{Cluster RV Evaluations}}{\text{Cluster GP Evaluations}}$\;
        %$r_{RV:GP} \longleftarrow $\RVToGPEvaluationRatio{}\;
        %\BlankLine
        %$p_{RV} \longleftarrow 1 - 0.5\cdot\frac{r_{RV:GP}}{\theta_{RV:GP}}$\;
        %\BlankLine
        $p_{RV} \longleftarrow 1 - 0.5\cdot\frac{\text{\#RV evaluations}}{\text{\#GP evaluations}}$\;
        %\If{$\mathcal{U}(0,1) < p_{RV}$}{
        %    $\mathcal{O},\mathcal{E},\mathcal{R}\longleftarrow$\RVStep{$\mathcal{O}$, $\mathcal{R}$, $\mathcal{E}$}
        %}\Else{
        %    $\mathcal{O},\mathcal{E} \longleftarrow$\GPStep{$\mathcal{O}$, $\mathcal{P}$, \TakeRandom{$\mathcal{F}_{GP}$}, $\mathcal{E}$}
        %}

        \If{$\mathcal{U}(0,1) < p_{RV}$}{
            %$\mathcal{O}, \mathcal{R}\longleftarrow$\RVStep{$\mathcal{O}$, $\mathcal{R}$}
            $\mathcal{O} \longleftarrow$ \RVStep{$\mathcal{O}$}
        }\Else{
            $\mathcal{O}\longleftarrow$ \GPStep{$\mathcal{O}$, $\mathcal{P}$, \TakeRandom{$\mathcal{F}$}}
        }
    }
    %\For{$i\in [0,N-1]$}{
    %    $\mathcal{P}_{i},\mathcal{E} \longleftarrow$\GPFI{$\mathcal{O}_{i}$, $\mathcal{E}$}
    %}

    $\mathcal{O} \longleftarrow$ \ApplyForcedImprovements{$\mathcal{O}$}\;
    $\mathcal{P} \longleftarrow \mathcal{O}$\;
}

\caption{\textbf{GP-RV-GOMEA}\label{code:rvgpg}}
\end{algorithm}
\vspace{-1.3cm}
\subsubsection{Forced Improvements} After the main variation steps have been performed, forced improvements are performed as described in~\cite{Sijben2022,Arkadiy2021}. The real-valued RV-GOMEA steps also perform forced improvements, however, as opposed to the procedure described in~\cite{Bouter17SO}, the real-valued forced improvements in the proposed method are modified to take the discrete structure into account. This is done by ensuring that the donor used makes use of constant values and then interleaving the discrete and real-valued forced improvements steps, as is done for the main variation shown in~\cref{code:rvgpg}. 
Since improvements can be both of structural and real-valued nature, the number of generations without improvements needed before real-valued forced improvements are applied was decreased from the default of 100 in RV-GOMEA to 20 generations.

\vspace{-0.5cm}
\subsubsection{Real-valued Intron Handling}
Similar to how GP-GOMEA can have introns, it is possible for real-valued constants to not be used in the encoded expression. To avoid unnecessary evaluations, changes to unused constants thus are not evaluated and are simply accepted in line with the intron handling of GP-GOMEA. Note that as we use the full FOS for RV-GOMEA, any change to the constants of a solution that actively uses at least one constant still has to be evaluated.

In RV-GOMEA, the constant values used for the maximum-likelihood estimation are selected based on the solution fitness. However, with the presence of introns, some of the values used for this estimate possibly do not contribute to the fitness of an individual. To avoid introducing noise through these inactive values, we make RV-GOMEA \textquote{intron aware} by filtering out these intron values in all steps where constant values are used to guide optimization. When selection is performed, for each constant index, the top \(35\%\) of the \emph{active values} are selected. Similarly, AMS is performed only on individuals with active constants and AVS only considers active values when updating the variance scaling factors.

\section{Experiments and Results\label{sec:experiments}}

In this section, we perform two types of experiments. The first experiment is performed using noise-free synthetic problems to isolate constant optimization and validate the effectiveness of the proposed approach. The second experiment uses real-world data to confirm the practical usefulness of the approach. %Furthermore, the SRBench\cite{SRBench2021} benchmark is used to compare the effectiveness of GP-RV-GOMEA to other state-of-the-art methods.

\subsection{Experimental Setup}

We compare the proposed approach without and with intron-aware (IA) model updates to GP-GOMEA with ERCs (\cite{virgolin2021improving}) and GP-GOMEA with coefficient mutation (\cite{CoefficientMutatioVirgolin2022}), hereafter abbreviated as RV, RV+IA, ERCs, and ERCs+CM, respectively. To isolate the effect of constant optimization from other techniques commonly used to increase performance such as linear scaling~\cite{Keijzer2004} (hereafter LS), constant tuning after optimization, and restarts, all combinations are tested. The other parameters used are detailed in~\cref{tab:parameters}. The function set was chosen based on~\cite{Nicolau2020}, where it was shown that this function set tends to generalize well to unseen data. %Other than the SRBench experiment
In all experiments, cross-validation with 5 folds and 7 repeats per fold using different seeds is used, corresponding to 35 runs per problem, method, restart, and LS configuration.

\begin{table}[b!]
\vspace{-0.5cm}
\centering
\caption{The parameter settings used in the experiments.\label{tab:parameters}}
\vspace{-0.25cm}
\small
\begin{tabularx}{0.95\linewidth}{>{\hsize=1.5\hsize\linewidth=\hsize}X|>{\hsize=0.875\hsize\linewidth=\hsize\centering\arraybackslash}X|>{\hsize=0.875\hsize\linewidth=\hsize\centering\arraybackslash}X|>{\hsize=0.875\hsize\linewidth=\hsize\centering\arraybackslash}X|>{\hsize=0.875\hsize\linewidth=\hsize\centering\arraybackslash}X}
\toprule
& \multicolumn{4}{c}{\thead{\textbf{Method}}} \\
\cline{2-5}
\textbf{Parameter} &  \thead{ERCs%\cite{virgolin2021improving}%
} & \thead{ERCs+CM%\cite{CoefficientMutatioVirgolin2022}%
} & \thead{RV} &  \thead{RV+IA} \\
\midrule
Objectives &  \multicolumn{4}{c}{Mean squared error (MSE)} \\
Tree height & \multicolumn{4}{c}{5\text{ (31 Nodes)}} \\
Operators &  \multicolumn{4}{c}{$+,-,\times,\div,\sin$} \\
Constant initialization &  \multicolumn{4}{c}{$\mathcal{U}(\min\{y_{train}\}, \max\{y_{train}\})$} \\
Termination &  \multicolumn{4}{c}{$10^7$ evaluations or convergence} \\
% Discrete Forced improvements NIS &  $1 + \log_{10}(\text{Population Size})$ \\
% \midrule
Population size & \multicolumn{4}{c}{$1000$} \\
Constant probability & \multicolumn{4}{c}{$50\%$} \\
%\midrule
%Coefficient mutation & & as per~\cite{CoefficientMutatioVirgolin2022} & & \\%\multicolumn{2}{c}{No} \\
% \midrule
% \multicolumn{5}{c}{GP-RV-GOMEA Specific} \\
\midrule
Number of constants & & & $10$ & $10$ \\
% $\theta_{RV:GP}$ & & & \multicolumn{2}{c}{$1$} \\
Intron Awareness & & & No & Yes \\
% Linear Scaling (LS) &  \(\{\text{Yes},\text{No}\}\) \\
\bottomrule
\end{tabularx}
%\vspace{-0.5cm}
\end{table}
% This was a bad idea
%\input{tables/syn}
We compare based on a fixed evaluation budget, as the different methods do not use the same number of fitness evaluations per generation. In addition, a run without restarts is stopped when it converges. A run is considered as converged when either all individuals encode the same structure or no evaluations were spent during a full generation. For the synthetic problems, a target mean squared error (MSE) value of \(10^{-8}\) was used as an additional termination criterion in the first experiment.  With restarts, the previous convergence conditions or no improvement to the elitist solution of the current restart within 10 generations trigger a full restart. Note that two elitist solutions are maintained, one to maintain the best solution across all restarts and one for the best solution of the current restart. A budget of \(10^7\) evaluations is used to ensure that all methods can converge within the computational budget. Hence, without restarts the fitness after convergence is compared, albeit this is not a fair comparison based on the actual number of evaluations spent. With restarts, the comparison based on evaluations is fair, however, the number of restarts or generations performed is not. Constant optimization after GP is performed using the L-BFGS implementation from PyTorch\cite{PyTorch2019} with a limit of 500 iterations, after which the resulting model is simplified using SymPy\cite{SymPy2017}.% The results after this post-processing (PP) step are marked as PP, and results before this PP step are marked as GP.

\subsection{Synthetic Problems: Does RV within GP Work as Desired?}

In the first experiment, we use the following synthetic problems with 1000 instances sampled with \(x_i\sim\mathcal{U}(-10,10)\):
\vspace{-0.3cm}
\begin{multicols}{2}
\begin{itemize}
\item[$\bullet$] \(-4.2\cdot x_0+\sqrt{2}x_1+e\cdot x_2\)
\item[$\bullet$] \(0.1\cdot x_0+0.2\cdot x_1+2.4\cdot x_2\)
\item[$\bullet$] \(\sin(\pi\cdot x_0)/(\pi \cdot x_0)\) %\(\frac{\sin(\pi x_0)}{\pi x_0}\)
\item[$\bullet$] \(\sin(1.772\cdot x_0) + \sin(2.035\cdot x_2)\)
\item[$\bullet$] \(\sin(\frac{\pi}{2}\cdot x_0 + \frac{\pi}{3}\cdot x_1)\)
\item[$\bullet$] \(\sin(\pi\cdot x + e)\)
\end{itemize}
\end{multicols}
\vspace{-0.3cm}

\begin{figure}[htbp]
    \vspace{-0.75cm}
    \centering
    %\fbox{%
    \includegraphics[width=0.85\linewidth]{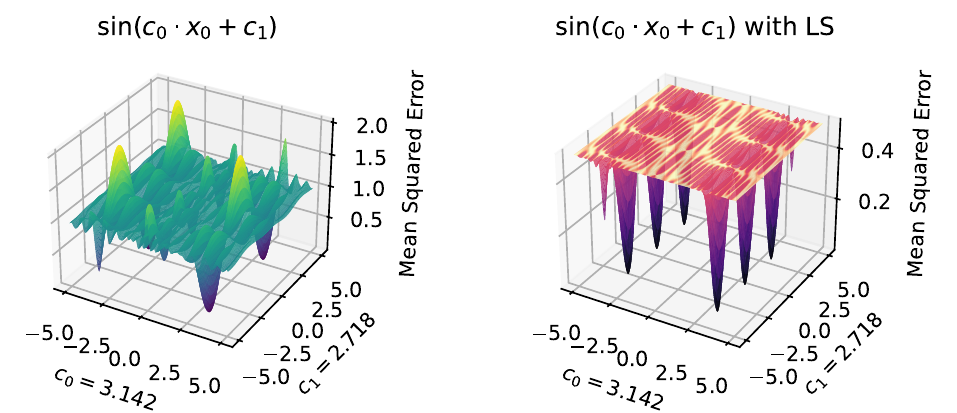}%}
    \vspace{-0.25cm}
    \caption{The constant optimization landscape can be multi-modal, both with and without linear scaling (LS). The error for every constant combination was computed using (the same) 500 instances sampled from \(\mathcal{U}(-10,10)\).\label{fig:landscape}}
\end{figure}

These problems were selected with specific criteria in mind. First, discovering the correct structure should not be overly challenging, to compare the constant optimization capabilities of various methods. Second, the chosen problems were designed to feature nested or non-linear combinations of constants, thereby ensuring that LS does not fully mitigate the need for constant optimization. Lastly, the synthetic problems were crafted to exhibit multi-modal constant optimization landscapes, motivating the use of gradient-free techniques. The landscape near the optimal constant values for \(\sin(\pi\cdot x + e)\) is shown in~\cref{fig:landscape}.

\begin{figure}[t]
    %\vspace{-0.5cm}
    \centering
    \includegraphics[width=\linewidth]{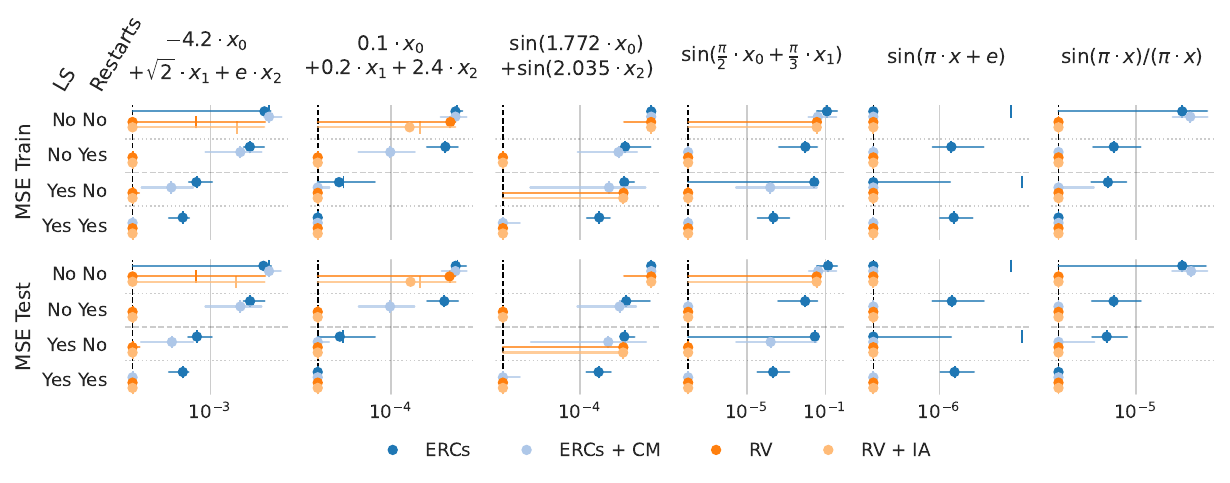}
    \vspace{-1cm}
    \caption{The training and testing MSE scores for the synthetic problems on a logarithmic scale. The bar corresponds to the median MSE before tuning, while the circle and horizontal line correspond
to the median and interquartile range (IQR) after post-processing. Note that the MSE was capped at the target value of \(1e-8\), which is highlighted with a vertical line.\label{fig:syn:mse}}
    \vspace{-0.7cm}
\end{figure}

The results are shown in~\cref{fig:syn:mse,tab:syn:success}. Note that this is an inherently biased comparison, as ERCs do not have an inherent ability to change the value of constants other than recombining several constants to represent new values not sampled initially. Likewise, coefficient mutation is random by nature, and thus the likelihood of improving constants decreases as the constant values get closer to the closest local optima. Nonetheless, we can confirm that our approach (i.e., RV and RV + IA) effectively optimizes constants, and clearly is better at reaching the MSE target of \(10^{-8}\) compared to the other constant optimization types considered (\cref{fig:syn:mse}). %\todo{IA also needs fewer evaluations as the gaussian adapts faster since there is less noise, but I currently don't show that.}{
\cref{tab:syn:success} shows that with intron awareness the proportion of runs reaching the MSE target increases, indicating that the real-valued distributions estimated have a better fit.

\begin{table}[htbp]
    \centering
    \caption{Proportion of runs that reach the \(10^{-8}\) MSE target. The best values per setting are highlighted in bold and the percentage in parentheses corresponds to the contribution of constant optimization after GP. The colored triangles indicate statistically significantly better methods (\(p<0.05\)) as per Fisher's exact test~\cite{Fisher1922}.\label{tab:syn:success}}
    \vspace{-0.25cm}
    \definecolor{Es}{HTML}{1f77b4}
    \definecolor{EC}{HTML}{aec7e8}
    \definecolor{RV}{HTML}{ff7f0e}
    \definecolor{RVIA}{HTML}{ffbb78}
    \scriptsize
    \begin{tabularx}{0.95\linewidth}{cc|@{}
    *{4}{S[table-format=3.1,detect-weight]@{}@{}l@{}X}@{}
}
\toprule
\thead{LS} & \thead{Restarts} & \multicolumn{3}{c}{\thead{ERCs \textcolor{Es}{\(\blacktriangledown\)}}} & \multicolumn{3}{c}{\thead{ERCs + CM \textcolor{EC}{\(\blacktriangledown\)}}} & \multicolumn{3}{c}{\thead{RV \textcolor{RV}{\(\blacktriangledown\)}}} & \multicolumn{3}{c}{\thead{RV + IA \textcolor{RVIA}{\(\blacktriangledown\)}}} \\
\midrule
\multicolumn{14}{c}{$-4.2\cdot x_0 + \sqrt{2}\cdot x_1+ e\cdot x_2$}\\
\midrule
\multirow[c]{2}{*}{No} & No & 37.1&\% (+37.1\%) &\includegraphics[width=0.95\linewidth]{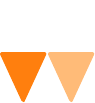}&2.9&\% (+0.0\%) &\includegraphics[width=0.95\linewidth]{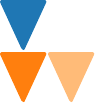}&\bfseries 60.0&\% (+25.7\%) &&\bfseries 60.0&\% (+22.9\%) &\\
& Yes & 0.0&\% (+0.0\%) &\includegraphics[width=0.95\linewidth]{images/FFTT.pdf}&5.7&\% (+0.0\%) &\includegraphics[width=0.95\linewidth]{images/FFTT.pdf}&88.6&\% (+0.0\%) &&\bfseries 97.1&\% (+0.0\%) &\\
\cline{2-14}
\multirow[c]{2}{*}{Yes} & No & 5.7&\% (+5.7\%) &\includegraphics[width=0.95\linewidth]{images/FFTT.pdf}&20.0&\% (+0.0\%) &\includegraphics[width=0.95\linewidth]{images/FFTT.pdf}&74.3&\% (+8.6\%) &&\bfseries 94.3&\% (+0.0\%) &\\
& Yes & 0.0&\% (+0.0\%) &\includegraphics[width=0.95\linewidth]{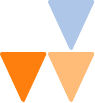}&74.3&\% (+0.0\%) &\includegraphics[width=0.95\linewidth]{images/FFTT.pdf}&\bfseries 97.1&\% (+0.0\%) &&94.3&\% (+0.0\%) &\\
\midrule
\multicolumn{14}{c}{$0.1\cdot x_{0}+0.2\cdot x_{1}+2.4\cdot x_{2}$}\\
\midrule
\multirow[c]{2}{*}{No} & No & 14.3&\% (+14.3\%) &\includegraphics[width=0.95\linewidth]{images/FFTT.pdf}&5.7&\% (+0.0\%) &\includegraphics[width=0.95\linewidth]{images/FFTT.pdf}&40.0&\% (+5.7\%) &&\bfseries 42.9&\% (+5.8\%) &\\
& Yes & 0.0&\% (+0.0\%) &\includegraphics[width=0.95\linewidth]{images/FTTT.pdf}&14.3&\% (+0.0\%) &\includegraphics[width=0.95\linewidth]{images/FFTT.pdf}&94.3&\% (+0.0\%) &&\bfseries 100.0&\% (+0.0\%) &\\
\cline{2-14}
\multirow[c]{2}{*}{Yes} & No & 34.3&\% (+2.9\%) &\includegraphics[width=0.95\linewidth]{images/FTTT.pdf}&71.4&\% (+0.0\%) &\includegraphics[width=0.95\linewidth]{images/FFTT.pdf}&97.1&\% (+0.0\%) &&\bfseries 100.0&\% (+0.0\%) &\\
& Yes & 97.1&\% (+0.0\%) &&\bfseries 100.0&\% (+0.0\%) &&\bfseries 100.0&\% (+0.0\%) &&\bfseries 100.0&\% (+0.0\%) &\\
\midrule
\multicolumn{14}{c}{$sin(1.772\cdot x_{0})+sin(2.035\cdot x_{2})$}\\
\midrule
\multirow[c]{2}{*}{No} & No & 2.9&\% (+2.9\%) &&5.7&\% (+0.0\%) &&\bfseries 14.3&\% (+2.9\%) &&\bfseries 14.3&\% (+0.0\%) &\\
& Yes & 0.0&\% (+0.0\%) &\includegraphics[width=0.95\linewidth]{images/FFTT.pdf}&11.4&\% (+0.0\%) &\includegraphics[width=0.95\linewidth]{images/FFTT.pdf}&82.9&\% (+0.0\%) &\includegraphics[width=0.95\linewidth]{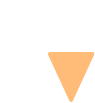}&\bfseries 100.0&\% (+0.0\%) &\\
\cline{2-14}
\multirow[c]{2}{*}{Yes} & No & 5.7&\% (+5.7\%) &\includegraphics[width=0.95\linewidth]{images/FTTT.pdf}&25.7&\% (+0.0\%) &&40.0&\% (+2.9\%) &&\bfseries 45.7&\% (+0.0\%) &\\
& Yes & 0.0&\% (+0.0\%) &\includegraphics[width=0.95\linewidth]{images/FTTT.pdf}&57.1&\% (+0.0\%) &\includegraphics[width=0.95\linewidth]{images/FFTT.pdf}&97.1&\% (+0.0\%) &&\bfseries 100.0&\% (+0.0\%) &\\
\midrule
\multicolumn{14}{c}{$sin(\frac{\pi}{2}\cdot x_0 + \frac{\pi}{3}\cdot x_1)$}\\
\midrule
\multirow[c]{2}{*}{No} & No & 14.3&\% (+14.3\%) &\includegraphics[width=0.95\linewidth]{images/FFTT.pdf}&0.0&\% (+0.0\%) &\includegraphics[width=0.95\linewidth]{images/TFTT.pdf}&34.3&\% (+0.0\%) &&\bfseries 45.7&\% (+0.0\%) &\\
 & Yes & 2.9&\% (+0.0\%) &\includegraphics[width=0.95\linewidth]{images/FTTT.pdf}&80.0&\% (+0.0\%) &\includegraphics[width=0.95\linewidth]{images/FFTT.pdf}&\bfseries 100.0&\% (+0.0\%) &&\bfseries 100.0&\% (+0.0\%) &\\
\cline{2-14}
\multirow[c]{2}{*}{Yes} & No & 28.6&\% (+28.6\%) &\includegraphics[width=0.95\linewidth]{images/FFTT.pdf}&20.0&\% (+0.0\%) &\includegraphics[width=0.95\linewidth]{images/FFTT.pdf}&\bfseries 85.7&\% (+0.0\%) &&80.0&\% (+0.0\%) &\\
  & Yes & 0.0&\% (+0.0\%) &\includegraphics[width=0.95\linewidth]{images/FTTT.pdf}&94.3&\% (+0.0\%) &&\bfseries 100.0&\% (+0.0\%) &&\bfseries 100.0&\% (+0.0\%) &\\
\midrule
\multicolumn{14}{c}{$sin(\pi\cdot x + e)$}\\
\midrule
\multirow[c]{2}{*}{No} & No & 85.7&\% (+85.7\%) &\includegraphics[width=0.95\linewidth]{images/FFFT.pdf}&74.3&\% (+0.0\%) &\includegraphics[width=0.95\linewidth]{images/FFTT.pdf}&97.1&\% (+0.0\%) &&\bfseries 100.0&\% (+0.0\%) &\\
& Yes & 0.0&\% (+0.0\%) &\includegraphics[width=0.95\linewidth]{images/FTTT.pdf}&\bfseries 100.0&\% (+0.0\%) &&\bfseries 100.0&\% (+0.0\%) &&\bfseries 100.0&\% (+0.0\%) &\\
\cline{2-14}
\multirow[c]{2}{*}{Yes} & No & 74.3&\% (+74.3\%) &\includegraphics[width=0.95\linewidth]{images/FFTT.pdf}&74.3&\% (+0.0\%) &\includegraphics[width=0.95\linewidth]{images/FFTT.pdf}&97.1&\% (+0.0\%) &&\bfseries 100.0&\% (+0.0\%) &\\
& Yes & 0.0&\% (+0.0\%) &\includegraphics[width=0.95\linewidth]{images/FTTT.pdf}&\bfseries 100.0&\% (+0.0\%) &&\bfseries 100.0&\% (+0.0\%) &&\bfseries 100.0&\% (+0.0\%) &\\
\midrule
\multicolumn{14}{c}{$sin(\pi\cdot  x)/(\pi\cdot x)$}\\
\midrule
\multirow[c]{2}{*}{No} & No & 40.0&\% (+40.0\%) &\includegraphics[width=0.95\linewidth]{images/FFTT.pdf}&2.9&\% (+0.0\%) &\includegraphics[width=0.95\linewidth]{images/TFTT.pdf}&\bfseries 88.6&\% (+0.0\%) &&80.0&\% (+0.0\%) &\\
& Yes & 5.7&\% (+0.0\%) &\includegraphics[width=0.95\linewidth]{images/FTTT.pdf}&\bfseries 100.0&\% (+0.0\%) &&\bfseries 100.0&\% (+0.0\%) &&\bfseries 100.0&\% (+0.0\%) &\\
\cline{2-14}
\multirow[c]{2}{*}{Yes} & No & 8.6&\% (+0.0\%) &\includegraphics[width=0.95\linewidth]{images/FTTT.pdf}&57.1&\% (+0.0\%) &\includegraphics[width=0.95\linewidth]{images/FFTT.pdf}&\bfseries 100.0&\% (+0.0\%) &&\bfseries 100.0&\% (+0.0\%) &\\
& Yes & 97.1&\% (+0.0\%) &&\bfseries 100.0&\% (+0.0\%) &&\bfseries 100.0&\% (+0.0\%) &&\bfseries 100.0&\% (+0.0\%) &\\
%\midrule
%A & & & & & F \\
\bottomrule
\end{tabularx}
\vspace{-0.5cm}
\end{table}

Both LS and restarts increase the average performance of all methods and decrease variance. With restarts, both RV variants reliably reach the target value in almost all runs. The constant optimization performed during post-processing is most noticeable for ERCs without LS or restarts, where it can lead to noticeable MSE improvements. However, compared to optimizing constants during optimization, the overall effect of tuning the best model after GP is small. Interestingly, tuning after and restarts have an unintuitive interaction when considering ERCs. Without restarts, the re-occurring subexpression \(\sin(\pi\cdot x)\) is often modeled with a constant that can be improved during post-processing. With restarts, however, many runs find \(\sin(x + x + x)\) instead, as \(3\cdot x \approx \pi\cdot x\). Since constant tuning was only applied before simplification, the lack of an explicit constant explains the decreased performance of ERCs with restarts.

\begin{figure}[htbp]
    %\vspace{-1cm}
    \centering
    \includegraphics[width=\linewidth]{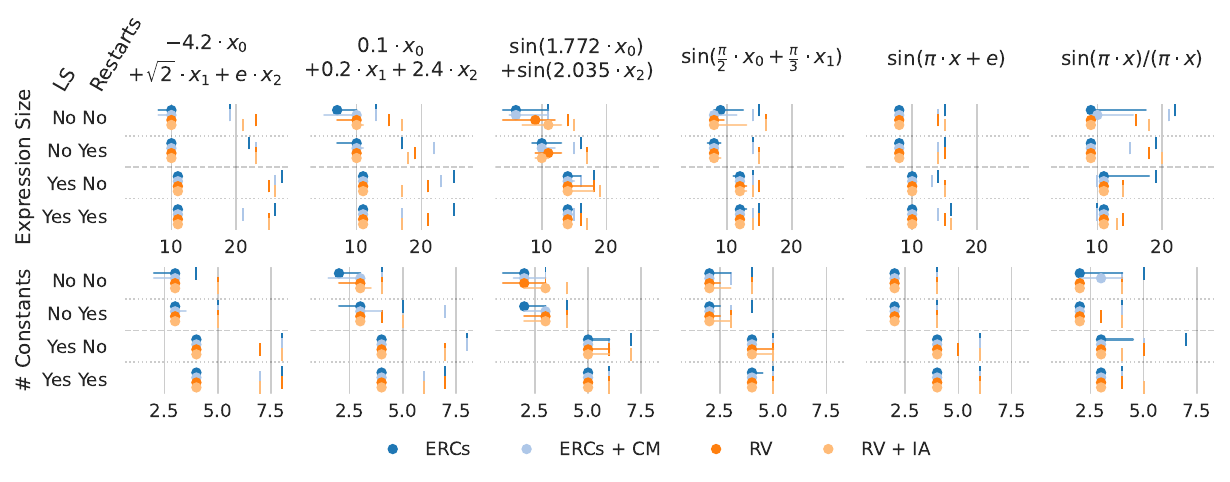}
    \vspace{-1cm}
    \caption{The number of constants used and expression sizes for the synthetic problems. The bar corresponds to the median before post-processing, while the circle and horizontal line correspond
to the median and IQR after post-processing.\label{fig:syn:expr}}
    %\vspace{-1cm}
\end{figure}

\begin{figure}[hbtp]
    \centering
    %\vspace{-0.75cm}
    \includegraphics[width=0.7\linewidth]{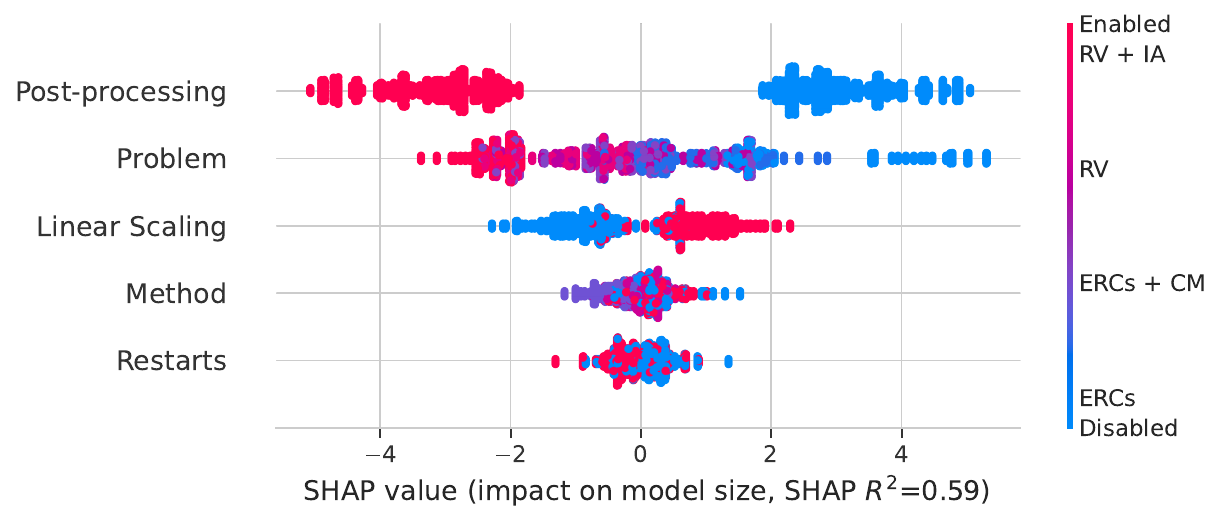}
    \vspace{-0.5cm}
    \caption{The SHAP\cite{SHAP2017} values show how different aspects influence expression size for the synthetic problems. The different methods are highlighted on the color map, for the other binary aspects the color indicates whether it was enabled or not.\label{fig:syn:shap:size}}
    \vspace{-0.5cm}
\end{figure}

In terms of the expressions found, in~\cref{fig:syn:expr,fig:syn:shap:size} we can see that apart from the simplification during post-processing, LS has a clear effect on both the number of constants used and the expression size, partly explained by the added scaling terms which add 2 constants and 4 nodes. The effect of simplification for the problems used tends to be considerable, for some problems the size after simplification is halved. The type of constant optimization used, however, has little impact on the expression size.

\subsection{Real-world Problems: GP-RV-GOMEA vs ERCs and Coefficient Mutation}

In this experiment, we consider the problems listed in~\cref{tab:problems} and compare the obtained expressions in terms of the coefficient of determination \(R^2\) score, expression size, and the number of constants used. Compared to the previous result, this experiment aims to provide a practically relevant comparison to the other forms of constant optimization.

\begin{table}
\centering
\vspace{-0.6cm}
\caption{The real-world problems used in the second experiment.\label{tab:problems}}
\vspace{-0.25cm}
\small
\begin{tabularx}{0.95\linewidth}{>{\hsize=1.75\hsize\linewidth=\hsize}X|>{\hsize=0.625\hsize\linewidth=\hsize\centering\arraybackslash}X|>{\hsize=0.625\hsize\linewidth=\hsize\centering\arraybackslash}X}
\toprule
\thead{Problem} & \thead{\# Instances} & \thead{\# Features} \\
\midrule
Airfoil Self-noise~\cite{DataAirfoil} &  1503 & 5 \\
Concrete Compressive Strength~\cite{DataConcrete} &  1030 & 8 \\
% Dow Chemical~\cite{?}&  1066 & 57 \\
Energy Cooling~\cite{DataEnergy} &  768 & 8 \\
Energy Heating~\cite{DataEnergy} &  768 & 8 \\
Yacht Hydrodynamics~\cite{DataYacht} &  308 & 6 \\
\bottomrule
\end{tabularx}
\vspace{-0.5cm}
\end{table}

\begin{figure}[h!]
    \vspace{-1cm}
    \centering
    \includegraphics[width=\linewidth]{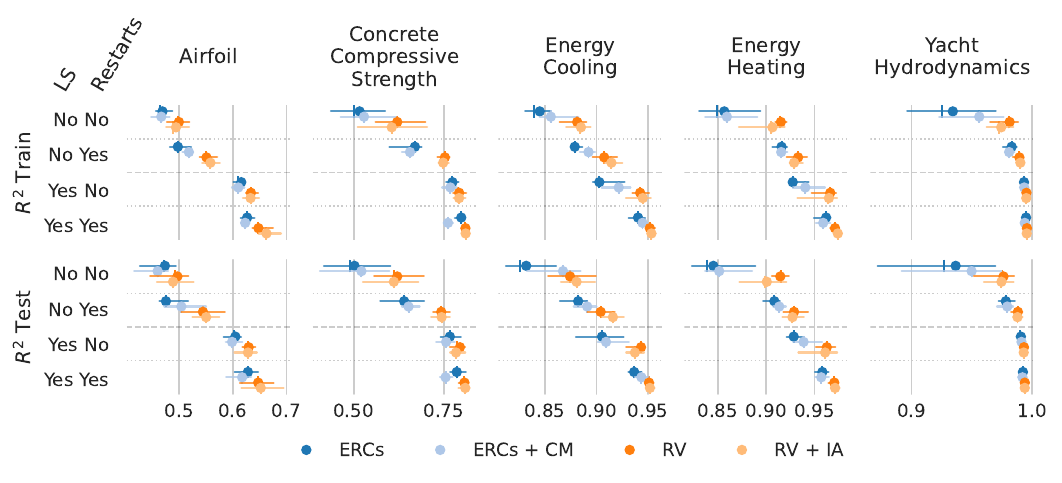}
    \vspace{-1cm}
    \caption{The training and testing \(R^2\)  scores for the real-world problems. The bar corresponds to the median \(R^2\) before tuning, while the circle and horizontal line correspond to the median and IQR after post-processing.\label{fig:bbo:r2}}
    %\vspace{-0.5cm}
\end{figure}

The results for the \(R^2\) scores displayed in~\cref{fig:bbo:r2} confirm that the proposed approach performs competitively on real-world problems, again outperforming both ERCs and coefficient mutation in terms of solution accuracy. Without LS or restarts, the RV version with intron-aware Gaussian model updates performs slightly worse compared to when intron awareness is not used. This could be explained by the noisy Gaussian distribution updates being more robust to changes in the expression structure between real-valued steps, as updating the distribution takes longer in contrast to the intron-aware version. With restarts or LS, however, the use of intron-awareness tends to perform a little better.

\begin{figure}[t!]
    %\vspace{-0.5cm}
    \centering
    \includegraphics[width=\linewidth]{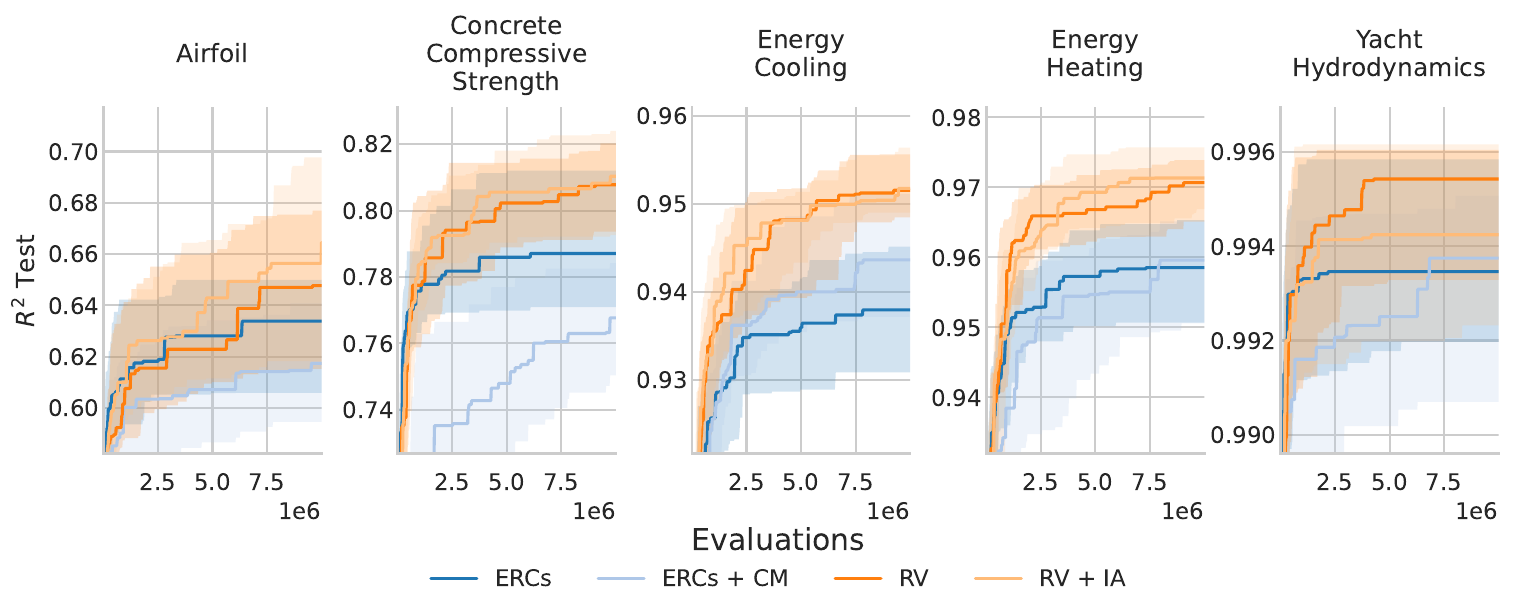}
    \vspace{-0.95cm}
    \caption{The median and IQR for the test \(R^2\) scores over evaluations spent with linear scaling and restarts enabled.}
    \label{fig:bbo:evals}
    \vspace{-0.7cm}
\end{figure}

\cref{fig:bbo:evals} shows that while ERCs have an initial advantage as all computational effort is spent on finding better structures, the importance of constant optimization becomes apparent as the evolution progresses. Improvements are still found close to reaching the computational budget, indicating that an increased budget could lead to improved results.

\begin{figure}[b!]
    \vspace{-0.8cm}
    \centering
    \includegraphics[width=\linewidth]{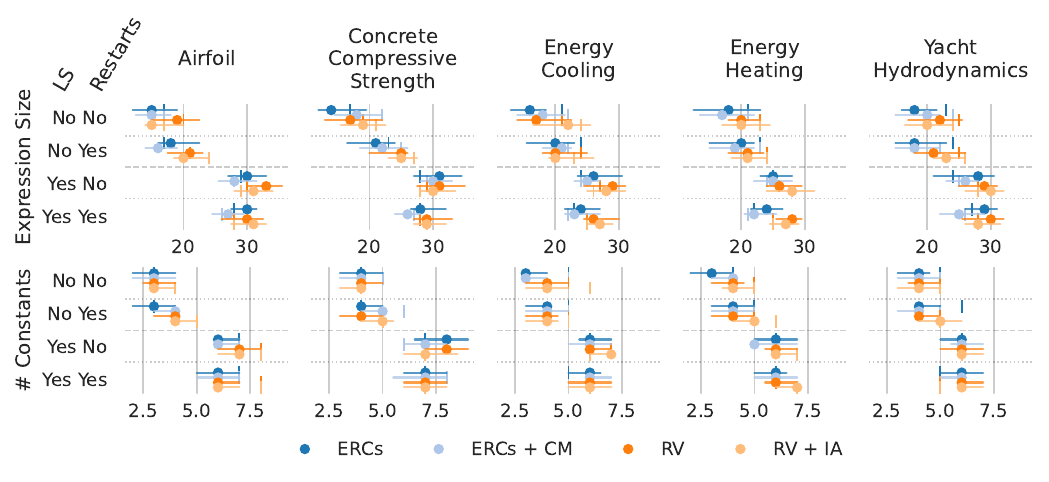}
    \vspace{-1cm}
    \caption{The number of constants used and expression sizes for the real-world problems. The bar corresponds to the median before post-processing, while the circle and horizontal line correspond
to the median and IQR after post-processing.\label{fig:bbo:expr}}
\end{figure}

Both LS and restarts improve accuracy and decrease variance for all methods tested, with LS having a bigger effect. In line with the synthetic problems in the previous experiment, the effect of tuning constants after GP is most noticeable for ERCs and in the absence of LS and restarts, but small compared to the effect of constant optimization during optimization. %The generalization gap between the training and testing accuracies is generally small, in line with~\cite{Nicolau2020}.

\begin{figure}[t!]
    \centering
    %\vspace{-0.75cm}
    \includegraphics[width=0.7\linewidth]{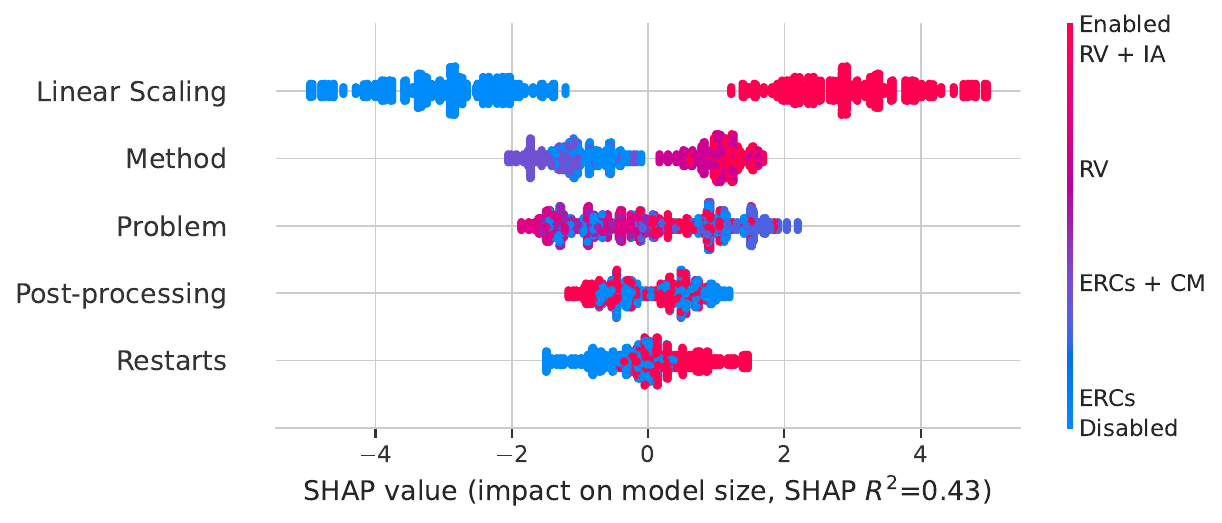}
    \vspace{-0.5cm}
    \caption{The SHAP\cite{SHAP2017} values show how different aspects influence expression size for the real-world problems. The different methods are highlighted on the color map, for the other binary aspects the color indicates whether it was enabled or not.\label{fig:bbo:shap:size}}
    \vspace{-0.5cm}
\end{figure}

In contrast to the synthetic problems,~\cref{fig:bbo:shap:size} indicates that next to LS, the RV constant optimization methods also lead to larger expressions while the effect of simplification decreased. This motivates a multi-objective approach.

Statistical testing results following the approach recommended by~\cite{Benavoli2017} are shown in~\cref{tab:bbo:st}, indicating that GP-RV-GOMEA is highly likely to lead to similar or noticeably better \(R^2\) scores compared to ERCs and coefficient mutation.

\begin{table}[hbtp]
    \vspace{-0.5cm}
    \centering
    \caption{The pair-wise probabilities of how likely \(M_1\) is to perform better, approximately equal (within \(0.01 R^2\)), or worse in terms of test \(R^2\) compared to \(M_2\) with LS and restarts, using a Bayesian hierarchical correlated t-test~\cite{Benavoli2017}.\label{tab:bbo:st}}
    \vspace{-0.25cm}
    \small
    \begin{tabularx}{0.95\linewidth}
    {XX|>{\raggedleft\arraybackslash}X>{\raggedleft\arraybackslash}X>{\raggedleft\arraybackslash}X}
        \toprule
        \thead{\(M_{1}\)} & \thead{\(M_{2}\)} & \thead{\(P(R^2_{M_{1}}\!\!>\! R^2_{M_{2}})\)} & \thead{\(P(R^2_{M_{1}}\!\!\approx\! R^2_{M_{2}})\)} & \thead{\(P(R^2_{M_{1}}\!\!<\! R^2_{M_{2}})\)} \\
        \midrule
        ERCs & ERCs + CM & 0.172 & 0.749 & 0.079 \\
        ERCs & RV & 0.019 & 0.400 & 0.581 \\
        ERCs & RV + IA & 0.025 & 0.205 & 0.771 \\
        \midrule
        ERCs + CM & RV & 0.059 & 0.114 & 0.827 \\
        ERCs + CM & RV + IA & 0.056 & 0.026 & 0.918 \\
        \midrule
        RV & RV + IA & 0.009 & 0.971 & 0.020 \\
        \bottomrule
    \end{tabularx}
    \vspace{-1cm}
\end{table}

\section{Discussion\label{sec:discussion}}

We proposed GP-RV-GOMEA, a new form of constant optimization in GP-GOMEA based on GOMEA-based mixed discrete and continuous optimization. Our experiments confirmed that simultaenous constant optimization across the whole population as opposed to per individual indeed works well for GP-GOMEA, and that our proposed approach clearly outperforms previous forms of constant optimization both with and without intron-aware real-valued Gaussian model updates. This holds for all combinations of linear scaling, restarts, and constant tuning after GP. Furthermore, on the problems considered, constant optimization during evolution has a positive effect on solution quality. However, on real-world problems, solution sizes tend to increase as well. While LS generally has the largest impact on accuracy, clearly, constant optimization too can have a noticeable impact.

The proposed approach introduces new parameters, such as the size of the constant pool available and the parameters of RV-GOMEA. These have not been explored extensively yet, as the goal of this work was to determine if such an approach is feasible and effective, revitalizing the research by~\cite{Howard1995}. Notably, not all mechanisms introduced in GAMBIT~\cite{Sadowski2018} were considered in this work, possibly increasing the effectiveness of this approach further.

Furthermore, the method has not yet been compared to on-line gradient-based constant optimization, which is a commonly used form of constant optimization in GP. Possibly a hybrid approach akin to basin hopping could prove to be more effective than only using one form of constant optimization.% While the proposed method has only been compared to GP-GOMEA, GP-GOMEA was compared to other methods in SRBench~\cite{SRBench2021}.

This work introduced a novel, intron-aware approach to updating the real-valued optimizer, showing increased performance on synthetic and real-world problems when combined with linear scaling or restarts. Notably, intron-awareness affects how fast the real-valued model can adapt to the changes in the real-valued fitness landscape introduced by structural changes. Compared to the intron-aware version, the decreased adaptivity can be seen as an implicit form of regularization, although no overfitting was found with the settings used.

While primarily caused by LS, the observed increases in expression size with constant optimization suggest that a multi-objective approach is needed to ensure small and accurate models.
Concerning constant tuning after GP, the observed interaction with restarts suggests that fine-tuning should be applied both before and after simplification to obtain better results.
%
%Overall, the proposed approach to optimize constants is effective and promising for future research.

\section{Conclusion\label{sec:conclusion}}
%Constant optimization has historically been an open issue in GP~\cite{ONeill2010}. While several approaches have been proposed over the years,
Over the years, several approaches to constant optimization in GP have been proposed, however, they come with drawbacks such as the need for gradients or large numbers of evaluations due to blind search. With this in mind, we proposed a novel, model-based, way of optimizing constants in GP-GOMEA, which we evaluated in the context of linear scaling, restarts, and optimization after GP on both synthetic and real-world data.

Our experiments confirm that optimizing constants across GP individuals can be effective and that simultaneous (evolutionary) constant tuning during GP can be required for increased performance. Compared to ERCs and coefficient mutation with the same underlying GP algorithm, we find that the proposed method improves overall expression accuracy in all settings considered, while achieving similar expression size.

\section{Acknowledgements}
This research was funded by the European Commission within the
HORIZON Programme (TRUST AI Project, Contract No.: 952060).

% ---- Bibliography ----
%
% BibTeX users should specify bibliography style 'splncs04'.
% References will then be sorted and formatted in the correct style.
\bibliographystyle{splncs04}
\bibliography{main}

\end{document}